\documentclass[10pt,twocolumn,letterpaper]{article}

\usepackage[pagenumbers]{iccv} 
\usepackage{multirow} 
\usepackage{float} 
\definecolor{iccvblue}{rgb}{0.21,0.49,0.74}
\usepackage[pagebackref,breaklinks,colorlinks,allcolors=iccvblue]{hyperref}

\def\paperID{12926} 
\def\confName{ICCV}
\def\confYear{2025}

\title{Document Haystack: A Long Context Multimodal Image/Document Understanding Vision LLM Benchmark}

\author{
\scalebox{0.88}{Goeric Huybrechts \hspace*{3mm}
Srikanth Ronanki \hspace*{3mm}
Sai Muralidhar Jayanthi \hspace*{3mm}
Jack Fitzgerald \hspace*{3mm}
Srinivasan Veeravanallur}\\
\scalebox{0.88}{\textit{Amazon AGI}}\\
\scalebox{0.88}{\{\texttt{huybrech,ronanks,saimucja,jgmf,srivee}\}@amazon.com}
}

\begin{document}
\maketitle
\vspace*{-3mm} 
\begin{abstract}
The proliferation of multimodal Large Language Models has significantly advanced the ability to analyze and understand complex data inputs from different modalities. However, the processing of long documents remains under-explored, largely due to a lack of suitable benchmarks. To address this, we introduce Document Haystack\footnote{\tiny Dataset: \url{https://huggingface.co/datasets/AmazonScience/document-haystack}}\footnote{\tiny Code: \url{https://github.com/amazon-science/document-haystack}}, a comprehensive benchmark designed to evaluate the performance of Vision Language Models (VLMs) on long, visually complex documents. Document Haystack features documents ranging from 5 to 200 pages and strategically inserts pure text or multimodal text+image ``needles" at various depths within the documents to challenge VLMs' retrieval capabilities. Comprising 400 document variants and a total of 8,250 questions, it is supported by an objective, automated evaluation framework. We detail the construction and characteristics of the Document Haystack dataset, present results from prominent VLMs and discuss potential research avenues in this area.
\end{abstract}

\vspace*{-3mm} 
\section{Introduction}
\label{sec:intro}

Large Language Models (LLMs) \cite{radford2018improving, chowdhery2023palm, touvron2023llama, touvron2023llama2} have fundamentally transformed natural language processing, demonstrating unprecedented capabilities in understanding and generating human-like text. This transformation has recently expanded with the emergence of multimodal LLMs, including GPT-4 \cite{achiam2023gpt}, Gemini 1.5 \cite{team2023gemini}, LLama 3.2 \cite{dubey2024llama}, Claude 3 \cite{claude3}, and Nova \cite{nova2024}. These advanced models excel at diverse tasks, from image captioning \cite{wang2024visionllm} to video reasoning \cite{jeoung2024adaptive} and speech translation \cite{mundnich2025zero}.

A particularly significant advancement has been in document understanding capabilities. This field is crucial across numerous domains, with particular importance in legal \cite{guha2023legalbench, cui2023chatlaw}, medical \cite{peng2023study, thirunavukarasu2023large}, and financial sectors \cite{li2023large, chen2024survey}, where accurate document interpretation directly impacts decision-making and regulatory compliance. The integration of LLMs has markedly improved key aspects of document processing, including information extraction accuracy, contextual understanding and question-answering capabilities based on document content.

However, document understanding presents distinct challenges that extend beyond conventional text processing. Documents frequently contain complex elements including tables, charts, and various visual components. These multilayered documents require sophisticated systems capable of processing diverse format types while efficiently managing large volumes of unstructured information.

While recent advances are promising, the effectiveness of Vision Language Models (VLMs) in processing multimodal documents remains an active area of research. Their capability to comprehend and analyze lengthy documents is not fully established as current evaluation frameworks focus on short documents, limiting our understanding of VLM performance on longer and more complex document analysis tasks.

In this work, we address this need by introducing Document Haystack, a novel benchmark designed to evaluate the performance of VLMs, in retrieving key multimodal information from long documents. The benchmark features 400 document variants ranging from 5 to 200 pages, with pure \textit{text} or multimodal \textit{text+image} needles inserted at different depths and in diverse formats. These needles, formatted as key-value pairs (e.g., ``The secret sport is “basketball”.", where “basketball” could be written as text or be illustrated as an image), are strategically placed to test the VLMs' ability to locate and extract specific information within extensive textual and visual contexts. Document Haystack provides a fully objective and comprehensive evaluation framework of 8,250 questions.

Results on our dataset indicate that VLMs achieve over 90\% accuracy in textual extraction from 200-page documents, but their performance drops by $\sim$30\% when the same documents are provided as images. Accuracy even falls to $\sim$40\% when retrieving combined text-image information, highlighting significant room for improvement in processing long visual documents.

\section{Related Work}
\label{sec:related_work}

The evaluation of LLMs has traditionally focused on short texts and specific tasks such as question answering, summarization, and translation, with early benchmarks like GLUE \cite{wang2018glue} and SuperGLUE \cite{wang2019superglue} playing instrumental roles in assessing natural language understanding capabilities. However, while these foundational benchmarks have been valuable, they primarily consist of short texts and tasks that do not require long-term context retention.

As LLMs advanced, the need for evaluating longer text comprehension became apparent, leading to the development of specialized long context benchmarks. Notable examples include the Needle in a Haystack \cite{kamradt2023needle}, which tests models' ability to retrieve information from extended contexts, and LongBench \cite{bai2023longbench}, which introduced the first bilingual, multi-task framework for assessing long-form text understanding.

The emergence of multimodal VLMs necessitated new evaluation approaches. Benchmarks such as VQA \cite{antol2015vqa}, NLVR \cite{suhr2018corpus}, and MileBench \cite{song2024milebench} were developed to assess models' capabilities in processing and reasoning about combined visual and textual information. Document understanding then emerged as a distinct challenge in multimodal evaluation. While early datasets focused on specific elements like charts \cite{masry2022chartqa} or single-page analysis \cite{mathew2021docvqa}, recent benchmarks such as DUDE \cite{van2023document}, Loong \cite{wang2024leave}, SlideVQA \cite{tanaka2023slidevqa}, and MMLongBench-Doc \cite{ma2024mmlongbench} attempted to tackle multi-page document comprehension.

Despite these advancements, there remains a significant gap in benchmarks that evaluate VLMs' performance on long, multimodal documents. Existing benchmarks either (1) lack the document length ($<$50 pages), (2) rely on independently pre-extracted text and images rather than original documents, or (3) don't allow a performance comparison of a similar task on different document lengths. While recently released benchmarks like MM-NIAH \cite{wang2024needle} and M-LongDoc  \cite{chia2024m} address some of these issues, they have their own constraints. MM-NIAH's limited prompt length ($<$72k tokens) makes it unsuitable for long document evaluation, while M-LongDoc, despite featuring documents spanning hundreds of pages, offers only 851 questions, employs a non-objective evaluation method, and doesn't support evaluating the same task with varying document lengths. Moreover, both rely on pre-extracting text and figures from the documents. This preprocessing approach prevents VLM providers from utilizing their native processing mechanisms on original PDF documents, which would provide a more accurate indication of their real-world performance. Although they are valuable contributions, there is a clear need for additional benchmarks. Document Haystack aims to fill this void.

\section{Benchmark Characteristics}
\label{sec:benchmark_characteristics}

Document Haystack is a comprehensive benchmark designed to evaluate the performance of VLMs in retrieving key multimodal information from long documents. The following sections detail the essential characteristics of our benchmark.

\begin{table}[h!]
\centering
\scalebox{0.6}{
\begin{tabular}{l|cccccccc|c}
\hline
\textbf{\# Pages}                               & \textbf{5} & \textbf{10} & \textbf{25} & \textbf{50} & \textbf{75} & \textbf{100} & \textbf{150} & \textbf{200} & \textbf{Total} \\ \hline
\multicolumn{10}{c}{\textbf{\textit{text} needles}}                                                                                                                                         \\ \hline
\textbf{\# Documents}                           & 25         & 25          & 25          & 25          & 25          & 25           & 25           & 25           & 200            \\
\textbf{\# Questions}                           & 125        & 250         & 625         & 625         & 625         & 625          & 625          & 625          & 4125           \\ \hline
\multicolumn{10}{c}{\textbf{\textit{text+image} needles}}                                                                                                                                        \\ \hline
\textbf{\# Documents}                           & 25         & 25          & 25          & 25          & 25          & 25           & 25           & 25           & 200            \\
\textbf{\# Questions}                           & 125        & 250         & 625         & 625         & 625         & 625          & 625          & 625          & 4125           \\ \hline
\multicolumn{10}{c}{\textbf{Total}}                                                                                                                                        \\ \hline
\textbf{Total \# Documents}  & 50         & 50          & 50          & 50          & 50          & 50           & 50           & 50           & \textbf{400}            \\
\textbf{Total \# Questions}                     & 250        & 500         & 1250        & 1250        & 1250        & 1250         & 1250         & 1250         & \textbf{8250}          \\ \hline
\end{tabular}
}
\caption{Document Haystack characteristics}
\label{tab:Characteristics}
\end{table}

\noindent\textbf{Needle in a Haystack} - Document Haystack extends the conceptual foundation of the Needle in a Haystack benchmark \cite{kamradt2023needle}, where language models are tasked with retrieving a specific statement (the ``needle") embedded within a long context window (the ``haystack").

\noindent\textbf{Document Selection and Lengths} - Document Haystack (Table \ref{tab:Characteristics}) comprises 25 publicly available financial 10-K reports, each originally exceeding 200 pages. These documents were selected to provide a realistic and challenging dataset for evaluating VLMs. To create varied testing conditions, each report was trimmed to different sizes: 5, 10, 25, 50, 75, 100, 150, and 200 pages. This variation allows for assessing the VLMs' performance across different context lengths. In total, our benchmark encompasses 400 document variants. 

\noindent\textbf{Needle Design} - Document Haystack consists of two parallel benchmark sets, distinguished by their different needle design (\textit{text} needles vs \textit{text+image} needles) as to evaluate different aspects of VLMs' capabilities. These parallel sets maintain identical document structures and distributions while varying only in how their target information is presented, enabling direct comparisons of VLMs' performance across different modalities.

\begin{figure}[h!]
  \includegraphics[width=\columnwidth]{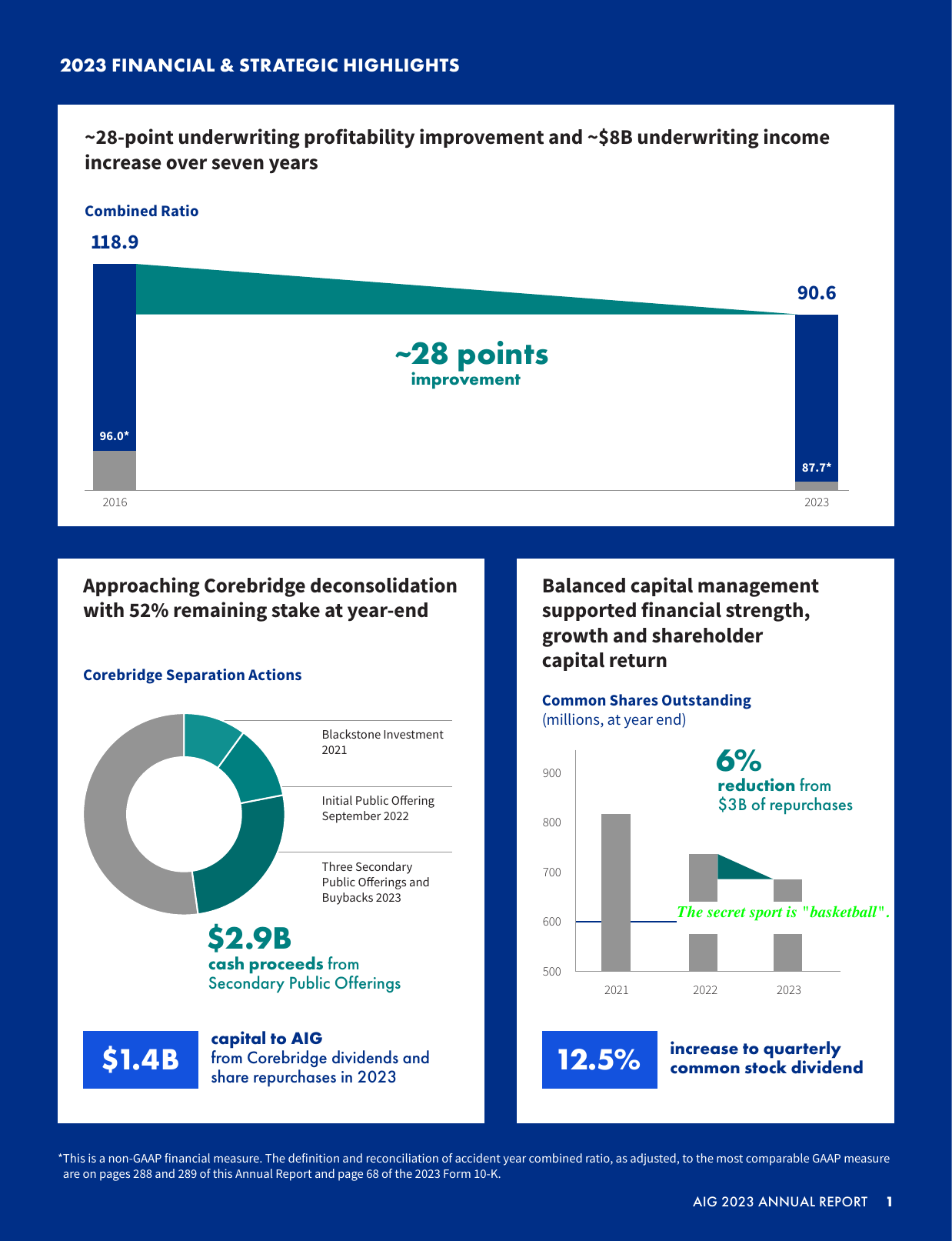}
  \caption{Example of a \textit{text} needle: Document page containing the hidden text-based information ``The secret sport is “basketball”.".}
  \label{fig:text_needle}
\end{figure}

\noindent\textbf{(1) \textit{text} needles} - The first set of our benchmark incorporates \textit{text} needles. To evaluate the VLMs' capability to extract specific text information from long documents, \textit{text} needles were inserted into each document as overlaid text. These needles are key-value pairs formatted as ``The secret KEY is “VALUE”.". For example, ``The secret sport is “basketball”." (Figure \ref{fig:text_needle}). The needles were inserted at various pages and spatial positions within the documents, with diverse colors, sizes, and fonts to add complexity.

\begin{figure}[h!]
  \includegraphics[width=\columnwidth]{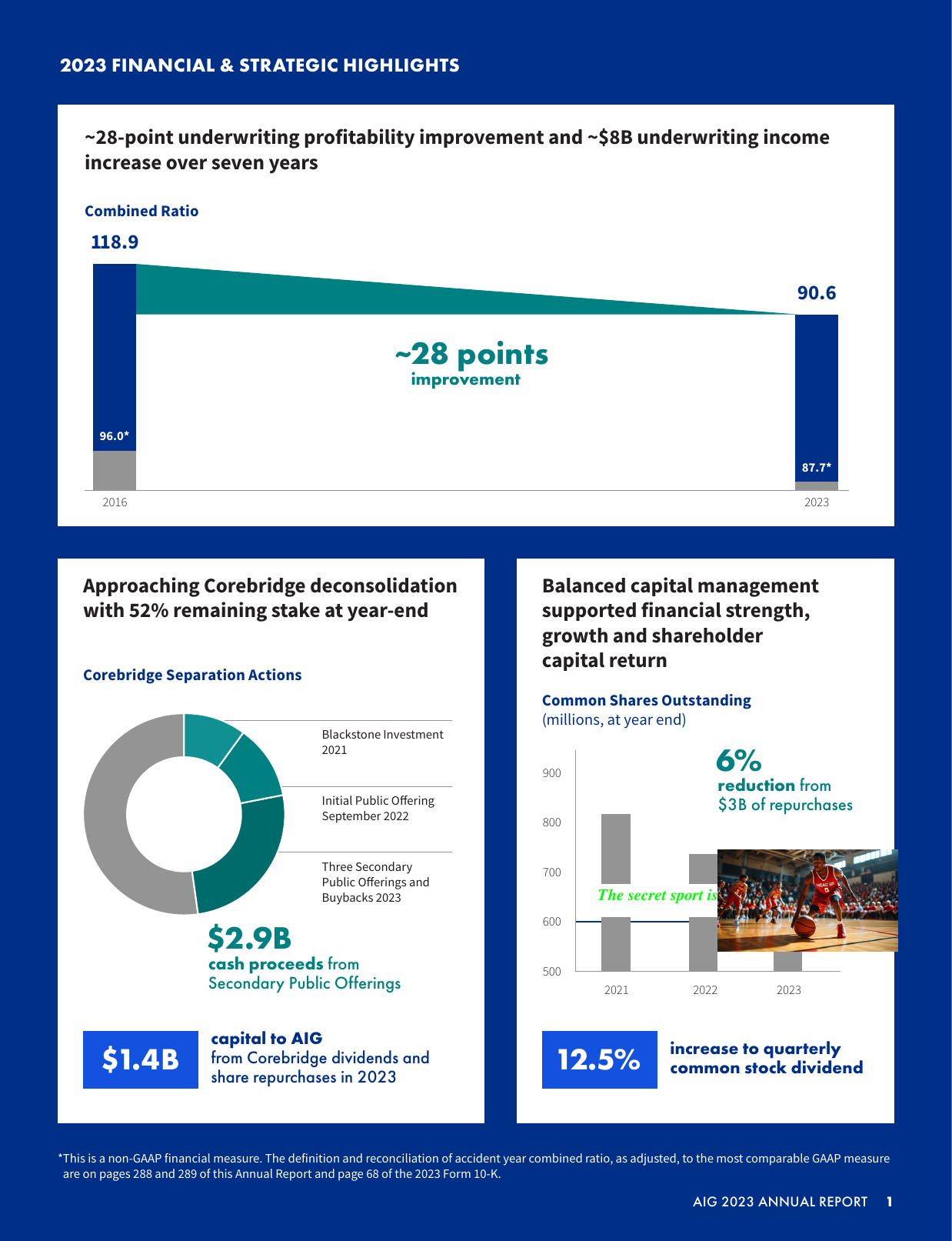}
  \caption{Example of a \textit{text+image} needle: Document page containing the hidden multimodal information where the text ``The secret sport is" is paired with an image of basketball.}
  \label{fig:image_needle}
\end{figure}

\noindent\textbf{(2) \textit{text+image} needles} - The second set of our benchmark introduces \textit{text+image} needles, maintaining identical positioning and typographical variations as the \textit{text} needles. The key distinction lies in the representation of ``VALUE" as an image rather than text (Figure \ref{fig:image_needle}). This modification enables assessment of VLMs' multimodal comprehension capabilities, requiring both textual and visual understanding for accurate information retrieval.

\noindent\textbf{Multiple Formats} - Document Haystack is available in three distinct formats (Table \ref{tab:formats}). First, both the \textit{text} and \textit{text+image} needle sets are provided in their original PDF format, preserving the complete document structure and formatting. Second, acknowledging that many VLM providers do not accept PDFs as input, we offer image versions of all documents, with each PDF page converted to a separate image file (200 DPI). Third, for scenarios requiring pure text processing, we provide a text-only version of the text needle set, extracted using the Python \textit{PdfReader}\footnote{https://pypdf.readthedocs.io/en/stable/modules/PdfReader.html\#the-pdfreader-class} text extraction tool. This multi-format approach enables researchers to evaluate VLMs or text-only LLMs under various input conditions while maintaining benchmark consistency.

\begin{table*}[h!]
\centering
\scalebox{0.85}{
\begin{tabular}{l|l|l|l}
\hline
\textbf{Benchmark Set} & \textbf{Format} & \textbf{Description} & \textbf{Use Case} \\
\hline
\multirow{3}{*}{(1) \textit{text} needles} 
    & PDF & Original document format & VLMs supporting PDF input \\
    & Image & 200 DPI page-wise images & VLMs requiring image input \\
    & Text & Extracted plain text & Text-only LLMs \\
\hline
\multirow{2}{*}{(2) \textit{text+image} needles} 
    & PDF & Original document format & VLMs supporting PDF input \\
    & Image & 200 DPI page-wise images & VLMs requiring image input \\
\hline
\end{tabular}
}
\caption{Document Haystack format variants}
\label{tab:formats}
\end{table*}

\noindent\textbf{Needle Distribution} - The distribution of needles was carefully designed to ensure they were placed at different depths within the documents. For documents with 5 and 10 pages, 5 and 10 needles were inserted, respectively. For documents with more than 25 pages, 25 needles were inserted. The pages for each needle were randomly selected from equal, non-overlapping page ranges. For instance, in a 100-page document, needle \#1 would be inserted on a randomly chosen page between 1 and 4, needle \#2 between pages 5 and 8, and so on.

\noindent\textbf{Needle Sets} - Five sets of 25 unique key-value pairs/needles were created (Table \ref{tab:key_values}), spanning diverse categories such as objects, animals, instruments, and fruits. Each set was allocated to five different reports, ensuring a diverse and comprehensive evaluation. This setup allows for a robust assessment of the VLMs' ability to locate and extract the correct values. Each key-value pair has an associated question. The presence of multiple needles within each document serves as intentional distractors to add complexity.

\begin{table}[h!]
\centering
\scalebox{0.65}{
\begin{tabular}{|c|l|l|}
\hline
\textbf{ID} & \textbf{Key} & \textbf{Values (Sets 1-5)} \\
\hline
1 & Shape & circle, triangle, rectangle, star, heart \\
2 & Currency & euro, dollar, pound, rupee, ruble \\
3 & Fruit & apple, banana, orange, grape, lemon \\
4 & Vegetable & carrot, broccoli, onion, mushroom, cauliflower \\
5 & Sport & basketball, tennis, boxing, skiing, surfing \\
6 & Food & pizza, hamburger, sausage, fries, chocolate \\
7 & Drink & coffee, tea, water, milk, smoothie \\
8 & Instrument & guitar, piano, trumpet, drum, violin \\
9 & Tool & hammer, wrench, saw, scissors, ruler \\
10 & Transportation & car, boat, train, airplane, bike \\
11 & Landmark & Eiffel Tower, Statue of Liberty, Taj Mahal,\\
  &  & Colosseum, Big Ben \\
12 & Kitchen appliance & blender, rice cooker, pan, toaster, microwave \\
13 & Flower & rose, sunflower, tulip, lavender, daisy \\
14 & Clothing & t-shirt, hat, glove, dress, sock \\
15 & Office supply & pencil, paperclip, stapler, envelope, calculator \\
16 & Animal \#1 & dog, cat, lion, elephant, giraffe \\
17 & Animal \#2 & zebra, kangaroo, panda, koala, penguin \\
18 & Animal \#3 & dolphin, shark, eagle, owl, spider \\
19 & Animal \#4 & snake, frog, turtle, horse, cow \\
20 & Animal \#5 & pig, bear, wolf, rabbit, squirrel \\
21 & Object \#1 & book, table, chair, door, clock \\
22 & Object \#2 & lamp, phone, key, watch, bottle \\
23 & Object \#3 & spoon, fork, knife, plate, bowl \\
24 & Object \#4 & umbrella, tree, bed, mirror, pillow \\
25 & Object \#5 & comb, toothbrush, towel, candle, vase \\
\hline
\end{tabular}
}
\caption{Five sets of 25 unique key-value pairs, where each value serves as a ``needle" that models must locate when prompted with its corresponding key.}
\label{tab:key_values}
\end{table}

\noindent\textbf{Animal and Object Categories} The inclusion of multiple animal and object sets (Animal \#1-\#5, Object \#1-\#5) serves a specific methodological purpose in evaluating VLMs' multimodal capabilities. While simpler categories like ``drink" or ``food" could potentially be answered by merely identifying relevant images in the document, animal and object categories require models to process both textual and visual information simultaneously. For instance, when multiple animal images appear in a document, the model must specifically identify which image corresponds to the ``secret Animal \#3" prompt, rather than simply detecting any animal image. This design ensures that VLMs truly combine their understanding of both modalities rather than relying solely on visual detection. These more challenging categories are exclusively used in documents of 25 pages or longer.

\noindent\textbf{Evaluation Methodology} - The purpose of Document Haystack is to retrieve the value associated with the secret key within the document. The evaluation question posed to the VLMs is: ``What is the secret KEY in the document?". To determine whether the VLM successfully found the needle, the response is lowercased and searched for the VALUE associated with the KEY. For image needles, we developed a comprehensive alias list to accommodate valid alternative responses (e.g., ``phone" vs. ``mobile," ``fries" vs. ``chips"). This straightforward approach ensures clear and objective assessment criteria across the benchmark's total of 8,250 questions.

\noindent\textbf{Objective} - The overarching goal of Document Haystack is to provide a comprehensive and challenging benchmark for evaluating the long context capabilities of multimodal LLMs, namely VLMs. By simulating real-world document processing scenarios and inserting specific information needles, the benchmark offers a rigorous test of an VLM's ability to accurately retrieve key multimodal information from lengthy and visually complex documents.

\noindent\textbf{Release} - The dataset and benchmarking code are publicly available at {\small\url{https://huggingface.co/datasets/AmazonScience/document-haystack}} and {\small\url{https://github.com/amazon-science/document-haystack}}.

\section{Results}
\label{sec:results}

\begin{table*}[h!]
\centering
\scalebox{0.9}{
\begin{tabular}{l|c|cccccccc}
\hline
\multirow{2}{*}{\textbf{Model}} & \multirow{2}{*}{\textbf{Avg. tokens/image}} & \multicolumn{8}{c}{\textbf{\#Pages}} \\
                                &                                              & \textbf{5} & \textbf{10} & \textbf{25} & \textbf{50} & \textbf{75} & \textbf{100} & \textbf{150} & \textbf{200} \\
\hline
Nova Lite & 1,626 & 8,130 & 16,260 & 40,650 & 81,300 & 121,950 & 162,600 & 243,900 & 325,200 \\
Gemini 2.0-Flash & 258 & 1,290 & 2,580 & 6,450 & 12,900 & 19,350 & 25,800 & 38,700 & 51,600 \\
GPT-4o-mini & 636 & 3,180 & 6,360 & 15,900 & 31,800 & - & - & - & - \\
\hline
\end{tabular}
}
\caption{Average image token consumption per model across different page lengths.}
\label{table:token_consumption}
\end{table*}

\begin{table*}[h!]
\centering
\begin{tabular}{l|cccccccc}
\hline
\multirow{2}{*}{\textbf{Model}} & \multicolumn{8}{c}{\textbf{\#Pages}}                                                                            \\
                                & \textbf{5} & \textbf{10} & \textbf{25} & \textbf{50} & \textbf{75} & \textbf{100} & \textbf{150} & \textbf{200} \\ \hline
Nova Lite              & \textbf{100.0}      & \textbf{98.8}        & 85.0        & 76.6        & \textbf{72.5}        & \textbf{69.6}         & \textbf{64.5}         & \textbf{62.9}         \\
Gemini Flash-2.0       & 83.2       & 74.8        & 82.7        & 64.0        & 63.2        & 58.4         &    46.9          &   51.8           \\
GPT-4o-mini            & 96.0       & 98.0        & \textbf{89.3}        &   \textbf{86.1}          & -           & -            & -            & -            \\ \hline
\end{tabular}
\caption{Accuracy on the \textbf{Text needles retrieval from document images} set - The task is to retrieve \textit{text} needles from documents provided a query (e.g., ``What is the secret fruit in the document?"). All documents are converted to images prior to VLM processing. The table shows the retrieval accuracy across different document lengths of three API providers: Nova Lite, Gemini Flash-2.0, and GPT-4o-mini.}
\label{tab:text_needles_as_image_performance}
\end{table*}

\begin{table*}[h!]
\centering
\begin{tabular}{l|cccccccc}
\hline
\multirow{2}{*}{\textbf{Model}} & \multicolumn{8}{c}{\textbf{\#Pages}}                                                                                                                                                  \\
                                & \textbf{5}           & \textbf{10}          & \textbf{25}          & \textbf{50}          & \textbf{75}          & \textbf{100}         & \textbf{150}         & \textbf{200}         \\ \hline
Nova Lite              & \textbf{100.0}      & \textbf{100.0}       & 98.9        & 95.2        & 94.6        & 93.9         & \textbf{94.1}         & 89.9         \\
Gemini Flash-2.0       & 99.2       & 99.6        & \textbf{99.5}        & 97.8        & \textbf{96.8}        & 97.1         & 91.5         & \textbf{91.8}         \\
GPT-4o-mini       & \textbf{100.0}       & \textbf{100.0}        & 97.9        & \textbf{98.4}        & 96.6        & \textbf{97.5}         & -         & -         \\ \hline
\end{tabular}
\caption{Accuracy on the \textbf{Text needles retrieval from parsed document text} set - The task is to retrieve \textit{text} needles from documents provided a query  (e.g., ``What is the secret landmark in the document?"). All documents are converted to text prior to VLM processing. The table shows the retrieval accuracy across different document lengths of three API providers: Nova Lite, Gemini Flash-2.0, and GPT-4o-mini.}
\label{tab:text_needles_as_text_performance}
\end{table*}

\begin{table*}[h!]
\centering
\begin{tabular}{l|cccccccc}
\hline
\multirow{2}{*}{\textbf{Model}} & \multicolumn{8}{c}{\textbf{\#Pages}}                                                                            \\
                                & \textbf{5} & \textbf{10} & \textbf{25} & \textbf{50} & \textbf{75} & \textbf{100} & \textbf{150} & \textbf{200} \\ \hline
Nova Lite              & \textbf{84.0}       & \textbf{84.0}        & 61.4        & 52.2        & 43.5        & 38.9         & 34.9         & 37.0         \\
Gemini Flash-2.0       & 53.6       & 52.0        & \textbf{67.4}        & \textbf{56.8}        & \textbf{48.6}        & \textbf{43.5}         & \textbf{37.9}         & \textbf{38.7}         \\
GPT-4o-mini            & 43.2       & 36.4        & 39.4        & 26.9       & -           & -            & -            & -            \\
\hline
\end{tabular}
\caption{Accuracy on the \textbf{Text+Image needles retrieval from document images} set - The task is to retrieve \textit{text+image} needles from documents provided a query (e.g., ``What is the secret flower in the document?"). All documents are converted to images prior to VLM processing. The table shows the retrieval accuracy across different document lengths of three API providers: Nova Lite, Gemini Flash-2.0, and GPT-4o-mini.}
\label{tab:image_needles_performance}
\end{table*}

In this section, we present the accuracy results from evaluating three prominent VLMs on our Document Haystack benchmark. We selected Nova, Gemini, and GPT as they are currently the only VLMs capable of processing more than 25 pages/images simultaneously. Other prominent VLMs have more limited image processing capabilities - for instance, Pixtral\footnote{https://docs.mistral.ai/capabilities/vision} \cite{agrawal2024pixtral} can only process 8 images at once, while Claude\footnote{https://docs.anthropic.com/en/docs/build-with-claude/vision}, despite providing support for 100 low-resolution images, is restricted to 20 images per API request for images larger than 2000x2000px.

Of these model families, we selected the following specific models for the benchmark evaluations: (1) Nova Lite, (2) Gemini Flash-2.0, and (3) GPT-4o-mini.

We report results across different formats (Table \ref{tab:formats}) of our benchmark to assess various capabilities of the VLMs. Specifically, we evaluate the models on following three sets:
\begin{enumerate}
    \item \textbf{Text needles retrieval from document images} (\ref{subsec:text_needles_as_images_results}): This set assesses the models' visual capabilities in retrieving textual information from documents presented as images.
    \item \textbf{Text needles retrieval from parsed document text} (\ref{subsec:text_needles_as text_results}): This set offers an additional comparison by evaluating the models' capability to retrieve the same \textit{text} needles from preprocessed, parsed document text.
    \item \textbf{Text+Image needles retrieval from document images} (\ref{subsec:text_image_needles_results}): This set assesses the models' visual capabilities in retrieving multimodal information from documents presented as images.
\end{enumerate}

\vspace*{-5mm}
We exclude PDF format results from our analysis due to limited direct PDF support among current VLM providers. However, as VLMs evolve to handle PDFs natively, this format presents an interesting future evaluation dimension. The PDF processing approaches of the various providers, whether converting documents to images, extracting text and visual elements separately, or employing hybrid methods, could significantly impact performance. Such comparisons would reveal not only the models' core capabilities but also the effectiveness of their document preprocessing pipelines, providing valuable insights into which architectural choices best preserve and utilize document structure and formatting. This additional evaluation axis could help guide the development of more sophisticated document processing strategies in future VLM implementations.

\subsection{Token Count}
\label{subsec:token_count}

Table \ref{table:token_consumption} presents the average token generation per image across the different VLMs, revealing significant variations in token consumption. Nova Lite generates approximately six times more tokens per image than Gemini (1,626 vs. 258 tokens). This difference becomes substantial when processing longer documents: for a 200-page document, Nova Lite processes 325,200 tokens compared to Gemini Flash-2.0's 51,600 tokens, while GPT-4o-mini theoretically falls in between at 127,200 tokens.

These token consumption patterns could significantly impact model performance, a topic that will be further explored in the following sections. Higher token counts, as seen in Nova Lite, may enable more detailed image information extraction but could potentially complicate specific information retrieval due to the longer context window that must be processed. Conversely, lower token counts, as demonstrated by Gemini, might offer more efficient processing but could miss subtle image details.

\subsection{Text Needles from Document Images}
\label{subsec:text_needles_as_images_results}

The performance metrics for the \textbf{Text needles retrieval from document images} set are presented in Table \ref{tab:text_needles_as_image_performance}. Our analysis reveals a consistent pattern across all models: accuracy deteriorates as document length increases. This degradation reaffirms the fundamental challenge in VLM processing: the inverse relationship between input context size and model performance.

Nova Lite and GPT-4o-mini demonstrate superior performance, achieving similar accuracy scores. However, GPT-4o-mini has a notable limitation: unlike the other models, it cannot process documents exceeding 50 pages. Nova Lite surpasses Gemini Flash-2.0 across all document lengths. Gemini's significantly lower token count per image, while potentially more efficient in terms of processing, may have inadvertently omitted crucial information necessary for effective information retrieval. Other model-specific factors such as memory management strategies may also contribute to these performance differences.

\begin{figure*}
    \centering
    \begin{subfigure}[b]{0.45\textwidth}
        \centering
        \includegraphics[width=\textwidth]{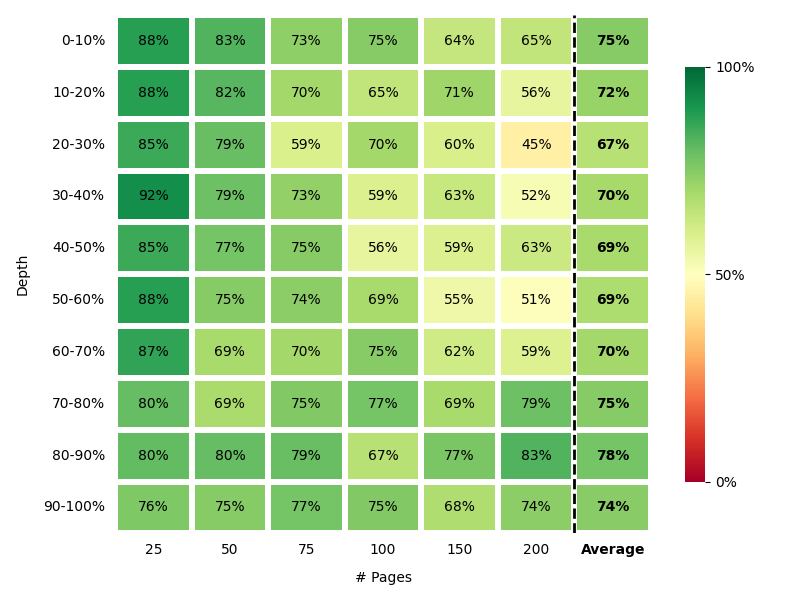}
        \caption{Nova Lite}
        \label{fig:NovaLite_TextNeedlesAsImages}
    \end{subfigure}
    \hfill
    \begin{subfigure}[b]{0.45\textwidth}
        \centering
        \includegraphics[width=\textwidth]{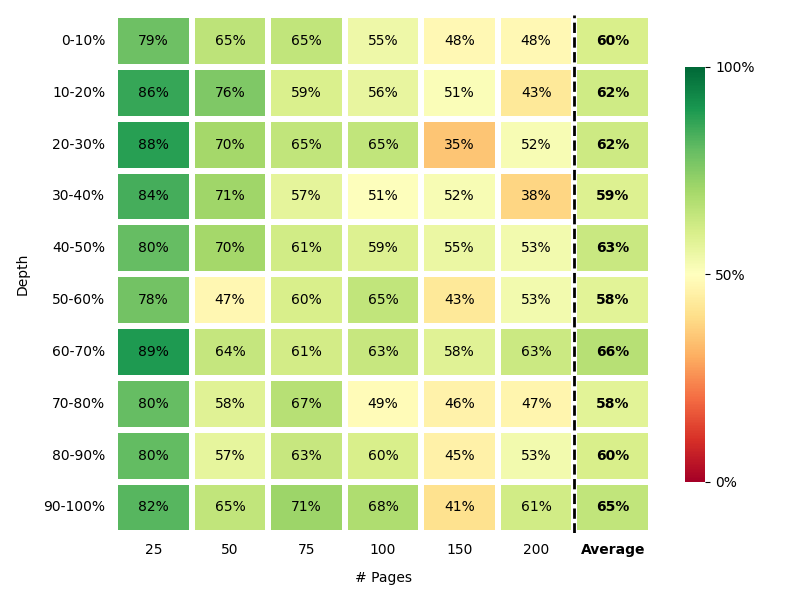}
        \caption{Gemini Flash-2.0}
        \label{fig:GeminiFlash_TextNeedlesAsImages}
    \end{subfigure}
    \caption{Performance analysis on the \textbf{Text needles retrieval from document images} set across document depths: Accuracy scores for different models evaluated over ten equidistant depth ranges, each containing approximately 63 needles.}
    \label{fig:depth_analysis_comparison_text}
\end{figure*}

Next, we report accuracy results across ten equidistant document depth ranges, each spanning 10\% of the document length. Due to our systematic needle distribution strategy, each depth range contains roughly a similar number of needles/questions ($\sim$63), enabling consistent comparison across depths. Since Nova and Gemini are the only models capable of processing documents up to 200 pages, we present detailed heatmap visualizations for these two models only in Figures \ref{fig:NovaLite_TextNeedlesAsImages} and \ref{fig:GeminiFlash_TextNeedlesAsImages}, respectively. While there are some slight variations, both models maintain relatively consistent performance across document depths.

\subsection{Text Needles from Parsed Document Text}
\label{subsec:text_needles_as text_results}

Table \ref{tab:text_needles_as_text_performance} presents performance metrics for the \textbf{Text needles retrieval from parsed document text} set, which establishes an upper bound for the \textit{Text needles retrieval from document images} set. The task of retrieving text information from pure text is inherently less challenging than extracting the same information from images, as it eliminates the complexity of visual processing \cite{wang2023cross, wang2024needle}.

The results demonstrate consistent performance across models and high accuracy across different context lengths, underscoring the models' remarkable capability to handle long context text inputs. It's worth noting that we encountered technical limitations with GPT-4o-mini for the 150 and 200-page sets, as these documents exceeded the API provider's token limit. Comparing these results with those in Table \ref{tab:text_needles_as_image_performance} reveals a significant performance gap between text-based and image-based text information retrieval. This difference illustrates the challenges and potential areas for advancement in understanding document text from images as opposed to pre-extracted text, highlighting the need for further research and development in this domain.

\subsection{Text+Image Needles from Document Images}
\label{subsec:text_image_needles_results}

Table \ref{tab:image_needles_performance} presents results from our most challenging benchmark set, \textbf{Text+Image needles retrieval from document images}, which reveals distinct performance patterns across document lengths. For shorter documents (5-10 pages), Nova Lite demonstrates remarkable superiority, outperforming other models by an absolute margin of 32+\%, while GPT-4o-mini shows the lowest accuracy. However, this performance hierarchy shifts with increasing document length: Gemini Flash-2.0 emerges as the leading model for longer documents, surpassing Nova Lite, albeit by a small margin.

This performance pattern yields an interesting insight regarding token consumption. While Gemini's lower token count appeared to disadvantage it in the \textit{Text needles retrieval from document images} set, it proves beneficial for \textit{image} needle extraction. Although differences in architecture, training, inference strategies, and other model aspects likely contribute, we hypothesize that this difference stems from the nature of the information being processed: \textit{text} needles in images require fine-grained detail preservation as they form a more subtle detail within the document page, whereas \textit{image} needles present more readily distinguishable visual elements. Consequently, Gemini's more compact token representation appears well-suited for image-based needle tasks, potentially contributing to its slight superior performance with longer documents.

\begin{figure*}
    \centering
    \begin{subfigure}[b]{0.45\textwidth}
        \centering
        \includegraphics[width=\textwidth]{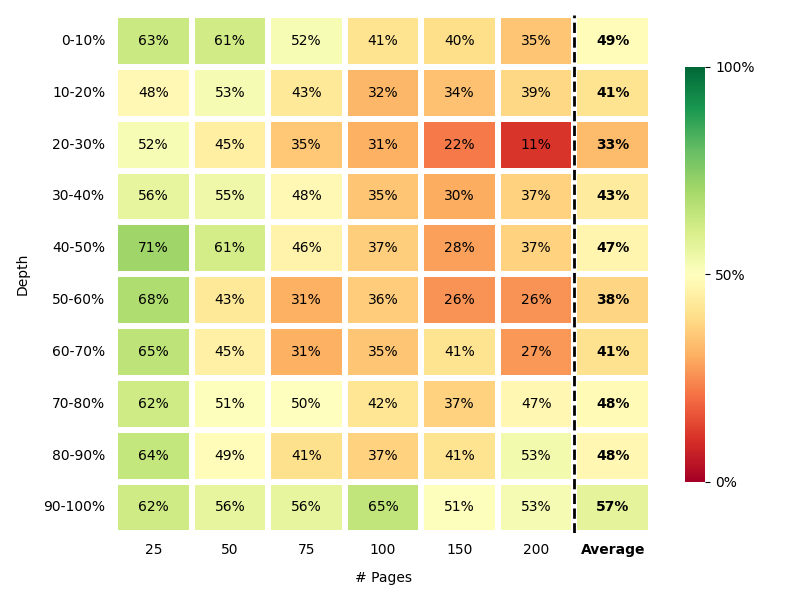}
        \caption{Nova Lite}
        \label{fig:NovaLite_ImageNeedles}
    \end{subfigure}
    \hfill
    \begin{subfigure}[b]{0.45\textwidth}
        \centering
        \includegraphics[width=\textwidth]{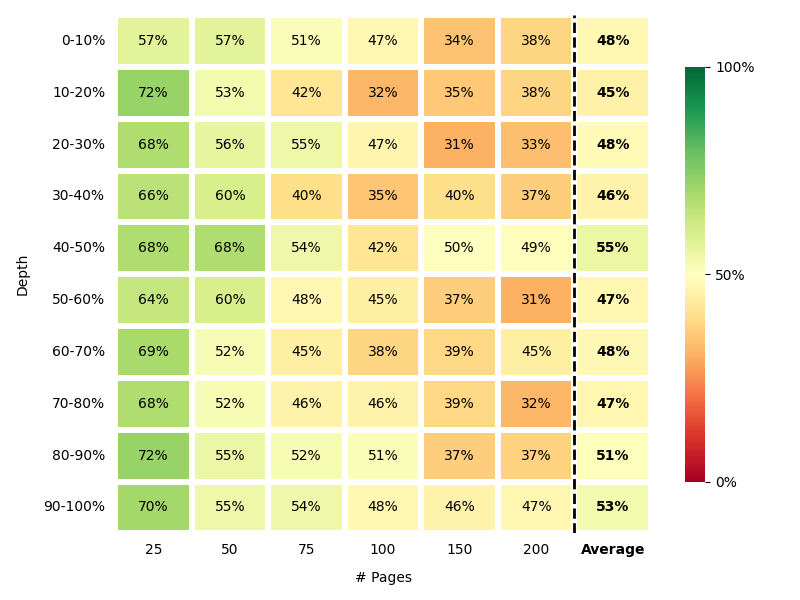}
        \caption{Gemini Flash-2.0}
        \label{fig:GeminiFlash_ImageNeedles}
    \end{subfigure}
    \caption{Performance analysis on the \textbf{Text+Image needles retrieval from document images} set across document depths: Accuracy scores for different models evaluated over ten equidistant depth ranges, each containing approximately 63 needles.}
    \label{fig:depth_analysis_comparison_text_images}
\end{figure*}

The heatmap analysis of our most challenging benchmark set reveals distinct patterns between models. Figures \ref{fig:NovaLite_ImageNeedles} and \ref{fig:GeminiFlash_ImageNeedles} present detailed performance visualizations for Nova Lite and Gemini Flash-2.0, respectively. Nova Lite exhibits a mild manifestation of the ``lost in the middle" phenomenon described in \cite{liu2024lost}, with slightly decreased accuracy in middle-depth ranges. In contrast, Gemini Flash-2.0 maintains consistent performance across all depth ranges, showing no discernible pattern of accuracy variation with document depth.

\section{Further Work}
\label{sec:further_work}

\subsection{Document Haystack Latency}

While our study emphasizes accuracy measurements, researchers could utilize our dataset to evaluate the latency characteristics of different VLMs in processing long documents. Such analysis would provide valuable insights into the computational efficiency and practical deployability of these models, particularly important for real-world applications where response time is crucial. This analysis could also reveal important trade-offs between model accuracy and response time, helping practitioners make informed decisions based on their specific application requirements.

\subsection{Advancing VLMs' Long Context Processing}

Our benchmark results reveal a significant disparity in VLMs' performance between textual and visual information processing over extended contexts. While models demonstrate robust capabilities in handling long context textual information, their performance in identifying and extracting image-based information deteriorates as document length increases. Additionally, the accuracy of extracting multimodal information is notably lower compared to extracting pure textual information from document images. These performance gaps highlight a critical area for improvement in next-generation VLMs.

The challenge likely stems from the increased complexity of maintaining and processing rich visual information over longer sequences, as opposed to textual information. Future research should focus on developing more efficient architectures, training approaches, and inference algorithms specifically designed to maintain visual context over extended sequences. This might include novel attention mechanisms, better visual tokenization strategies, or more sophisticated methods for managing the interaction between textual and visual modalities in long documents. Addressing these limitations is crucial as real-world applications often require processing lengthy documents containing both text and images, from technical manuals to medical records.

\subsection{Extended Analysis Framework}

While our primary contribution is a benchmark dataset for long context multimodal retrieval challenges, we recognize the importance of enabling more fine-grained analysis tasks. To facilitate this, our dataset release includes comprehensive metadata for each needle, comprising: (1) page location within the document, (2) precise X-Y coordinates on the page, (3) image size, (4) color specifications, and (5) font type and size.

This rich metadata will help support various research directions beyond basic retrieval, such as location-aware information extraction, spatial relationship analysis, visual attribute understanding and document layout comprehension. These additional data points provide researchers with the flexibility to design and evaluate more sophisticated tasks that reflect different real-world document processing challenges. We envision this expanded functionality will drive innovation in both model architectures and evaluation methodologies for long context multimodal understanding.
\section{Conclusion}
\label{sec:conclusion}

Document Haystack marks a significant advancement in the evaluation of Visual Language Models (VLMs) by providing a comprehensive benchmark for assessing their performance on long, visually complex documents. Our findings highlight some limitations of current models, underscoring the need for continued research and development. We believe that Document Haystack will serve as a valuable resource for the research community, driving innovation and progress in the field of multimodal document understanding and paving the way for more effective long context VLMs.
{
    \small
    \bibliographystyle{ieeenat_fullname}
    \bibliography{main}
}

\end{document}